\documentclass[conference]{IEEEtran}
\IEEEoverridecommandlockouts
\usepackage{cite}
\usepackage{amsmath,amssymb,amsfonts}
\usepackage{algorithmic}
\usepackage{graphicx}
\usepackage{textcomp}
\usepackage{xcolor}
\usepackage{hyperref}
\usepackage[ruled,vlined]{algorithm2e}
\hypersetup{
	colorlinks=true,
	linkcolor=blue,
    filecolor=magenta, 
    urlcolor=cyan
}
\def\BibTeX{{\rm B\kern-.05em{\sc i\kern-.025em b}\kern-.08em
    T\kern-.1667em\lower.7ex\hbox{E}\kern-.125emX}}
\begin{document}

\title{Image Processing Techniques for identifying tumors in an MRI image}

\author{\IEEEauthorblockN{Jacob John}
\IEEEauthorblockA{\textit{School of Computer Science and Engineering (SCOPE)} \\
\textit{Vellore Institute of Technology}\\
Vellore, India \\
jacob.john2016@vitalum.ac.in}
}

\maketitle

\begin{abstract}
Medical Resonance Imaging or MRI is a medical image processing technique that used radio waves to scan the body. It is a tomographic imaging technique, principally used in the field of radiology. With the advantage of being a painless diagnostic procedure, MRI allows medical personnel to illustrate clear pictures of the anatomy and the physiological processes occurring in the body, thus allowing early detection and treatment of diseases. These images, combined with image processing techniques may be used in the detection of tumors, difficult to identify with the naked eye. This digital assignment surveys the different image processing techniques used in Automated Tumor Detection (ATD). This assignment initiates the discussion with a comparison of traditional techniques such as Morphological Tools (MT) and Region Growing Technique (RGT). 
\end{abstract}


\section{Introduction}
MRIs provide us with the technology to detect tumors or neoplasms at an early stage and provides essential information for early disease detection, i.e., identify abnormal or diseased tissue. A neoplasm is this uncoordinated abnormal and excessive growth of a tissue occurring inside the body \cite{b1}. This growth is referred to as a tumor when it forms a mass. However, it should be noted that neoplasms do not always form a mass \cite{b2} and some do not form a tumor such as leukemia and forms of carcinoma in situ. Furthermore, the growth of a neoplasm is independent of its surrounding tissues. Regardless of removing the original growth trigger \cite{b3}, the neoplasm or tumor persists to grow at an abnormal rate \cite{b4}\cite{b5}. Thus, presenting a threat to the human anatomy. 

\subsection{Neoplasms or tumors}

Neoplasm can be grouped into five categories as according ICD-O behavior codes \cite{b6} and ICD-10 \cite{b7}. These include:
\begin{itemize}
\item{Benign neoplasms – noncancerous,}
\item{Neoplasms of uncertain and unknown behavior,}
\item{Carcinoma in situ – These will not spread and grow in situ. These could potentially be cancer \cite{b8}.}
\item{Malignant neoplasms stated or presumed to be primary, of lymphoid, hematopoietic and related tissue and}
\item{Malignant neoplasms of ill-defined, secondary and unspecified sites.}
\end{itemize}

A malignant neoplasm or malignant tumor is also known as cancer. These cells divide and grow excessively to form lumps that are cancerous \cite{b9}. Hence, spreading to other parts of the body and invading healthy tissues. Treatments include chemotherapy and radiation therapy that are used to kill cancer cells throughout and specific parts of the body, respectively.

\subsection{Magnetic Resonance Imaging}

\begin{figure}[htbp]
\centerline{\includegraphics[width=0.4\textwidth]{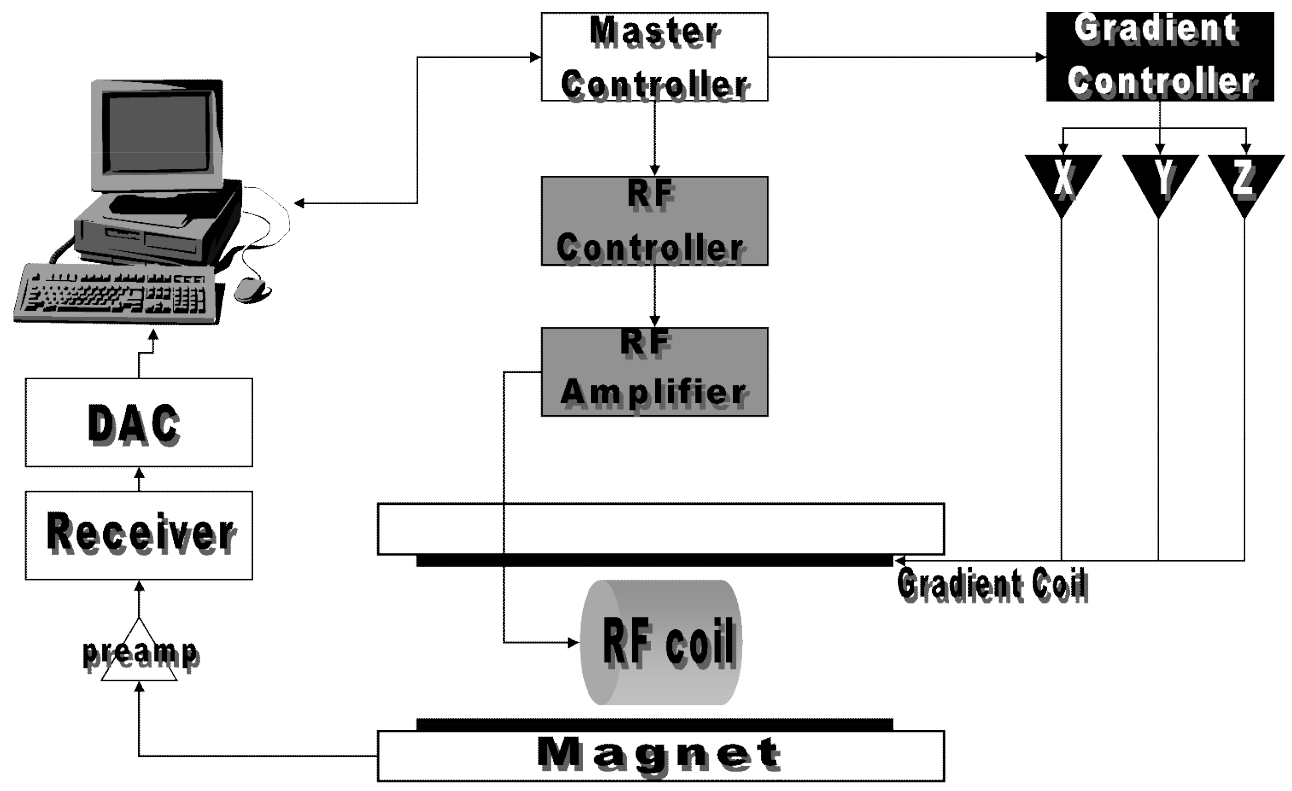}}
\caption{A Magnetic Resonance Imaging System (MRI), simplified. \cite{b11}}
\label{fig1}
\end{figure}

Rather than using X-rays or ionizing radiation like CAT or PET scans do, MRI scanners use radio waves and strong magnetic fields to produce cross-sectional images of the internal anatomy of the body. An MRI system works on the principle of nuclear magnetic resonance (NMR) and consists of the following components as depicted in figure \ref{fig1}
\begin{itemize}
\item{The main magnet used to generate a strong uniform static field or the $B_0$ field. This partially polarizes the nuclear spins and causes the hydrogen atom to line up in the direction of the field. The strength of the magnetic field produced by this magnet is typically between 0.5 tesla to 2.0 tesla \cite{b10}}
\item{The magnetic field gradient system consisting of a gradient controller and a gradient coil.}
\item{The radio frequency (RF) system consists of an RF coil, an RF amplifier, and an RF controller. The RF transmitter coil generates a rotating magnetic field, $B_1$, for exciting a spin system in the unmatched protons. This specific resonance frequency is based on the tissue being image and is termed as the Larmour frequency.}
\item{The receiver coil is connected to the computer system via a Digital to Analog converter (DAC). This coil converts the magnetization into an electric signal for imaging. }
\end{itemize}

\subsection{Fourier Transforms}

\begin{figure}[htbp]
\centerline{\includegraphics[width=0.4\textwidth]{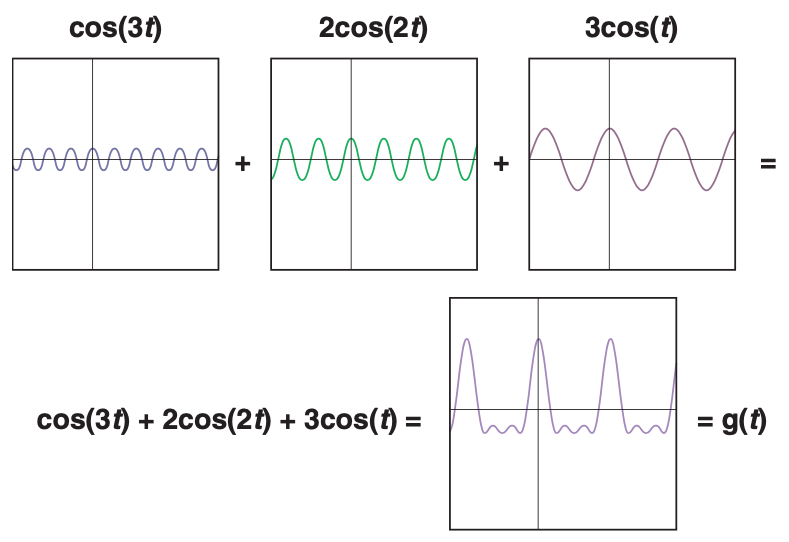}}
\caption{Generating a complicated signal by superimposing three simpler waves. \cite{b12}}
\label{fig2}
\end{figure}

The magnetic signal received is decomposed as the sum of a series of simple waves with varying amplitudes and frequencies using Fourier transforms (FTs) \cite{b12}. Figure \ref{fig2} illustrates this decomposition from a complicated signal to simple waves.  

FT isolates the critical components of an image such as by expressing the signal (i.e., a function of time) into its underlying frequencies. FT are classified as orthogonal sinusoidal basis function, and this is known as the frequency domain representation of the original signal. Equation \eqref{eq1} defines the frequency domain representation or Fourier transform of a continuous function of time, $f(t)$ \cite{b13}. While, equation \eqref{eq2} denotes the same equation using Euler’s formula, $e^{j\theta}=\cos\theta+j\sin\theta$.  Note that since $t$ is integrated out, we can rewrite $\mathfrak{I}\{f(t)\} $ as a function of $\mu$. We can represent this as $\mathfrak{I}\{f(t)\} = F(\mu)$.

\begin{equation}
\mathfrak{I}\{f(t)\} = \int_{-\infty}^{\infty} f(t)e^{-2\pi j\mu t}dt\label{eq1}
\end{equation}

\begin{equation}
F(\mu) = \int_{-\infty}^{\infty} f(t) [\cos(2\pi \mu t) - j \sin(2\pi \mu t)]dt\label{eq2}
\end{equation}

where, $t$ and $\mu$ are continuous variables.

\subsection{Image Acquisition in MRI}

\begin{figure}[htbp]
\centerline{\includegraphics[width=0.4\textwidth]{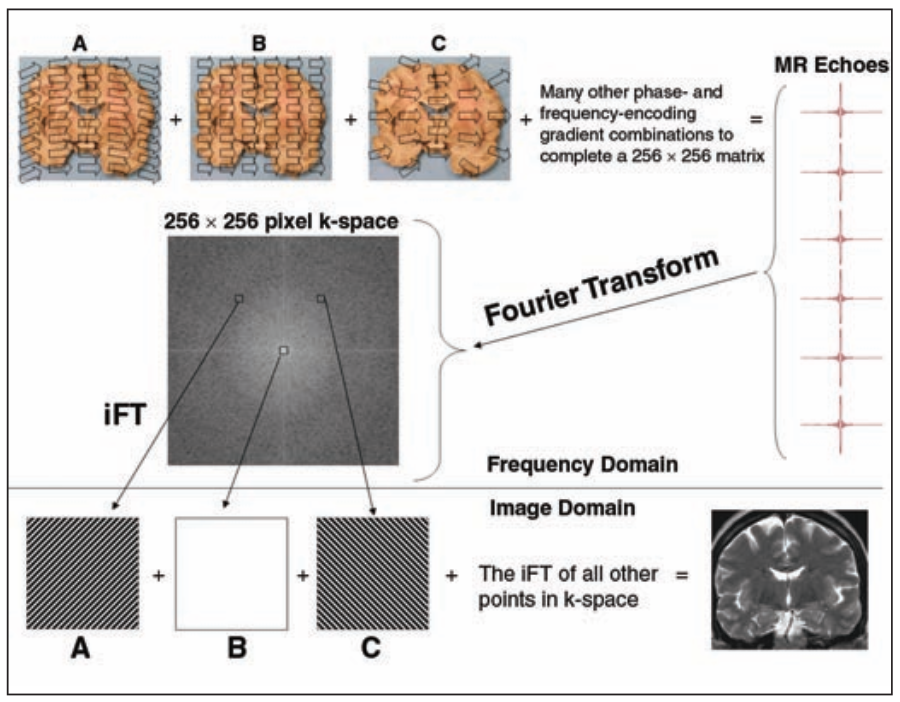}}
\caption{A small part of a coronal slice of a brain interrogated for all its spatial frequencies and amplitude information in Fourier k-spaces. The summation of relative frequencies and the IFT of all other points in k-space contributes to give the image space. \cite{b12}}
\label{fig3}
\end{figure}

\begin{figure}[htbp]
\centerline{\includegraphics[width=0.4\textwidth]{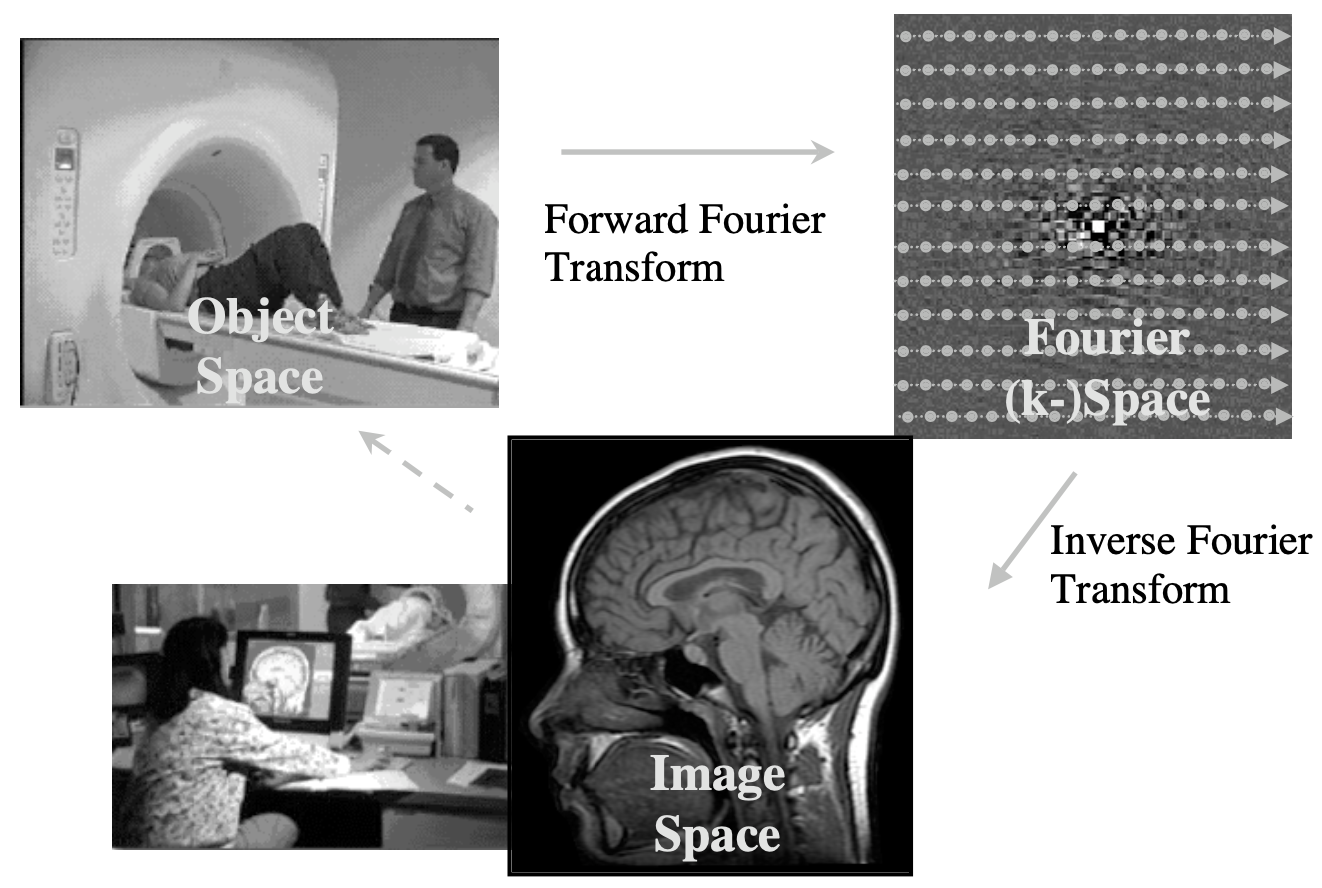}}
\caption{The MR imaging process used in image acquisition. A simplified overview. \cite{b11}}
\label{fig4}
\end{figure}

Since all the spin systems in the protons process at the same frequency and phase dictated by the magnetic field, $B_0$, a dynamically changing gradient field is applied for the separation of spin systems \cite{b14}. This is followed by applying the FT on the digitized signal and converting it into its Fourier k-space. A k-space is where the signal is organized into its spatial frequencies and amplitude information. Figure \ref{fig3} depicts this process. An inverse Fourier transform (IFT) is then applied to transform the image to the image space as shown in figure \ref{fig3}. This entire step by step process can be illustrated in a simplified manner by Figure \ref{fig4}. The figure gives us an overview of the MR imaging process from a signal processing perspective.

\section{Image Processing Techniques}

The previous section discussed the definition of a tumor and the process of MRI. Furthermore, it introduced the Fourier Transform and discussed its purpose in imaging the digitized signal received from the MRI system.
The following section will discuss the step by step techniques in detecting tumors in the images received from MRI scans. Following image acquisition, this section focuses on segmentation techniques such as MTs and RGTs in order to identify tumors at an early stage.

Patil and Bhalchandra present a MATLAB step by step implementation of brain tumor extraction in \cite{b15}. This method incorporates filters for noise removal, filters for enhancement, segmentation and morphological operations to detect the tumor.

\subsection{Preprocessing}

\subsubsection{Acquiring a grayscale MRI scan}

This is the first step of any image processing. The object of interest is captured by a sensor (e.g., camera) and then digitized using an analog to digital converter. 

The acquired magnetic resonance images are represented in grayscale.  The intensity or amplitude of a grayscale image is represented as a function $f(x,y)$ where $x$ and $y$ are the spatial coordinates of the image. It should be noted that $f,x$ and $y$ are finite and discrete quantities. Since a grayscale image is represented as an 8-bit image, the value ranges from 0 to 255. With 0 being the weakest intensity and represented as the color black. This is due to the absence of light. While, 1 being the strongest intensity and represented as the color white. This is caused by the ``total transmission of light at all visible wavelengths" \cite{b16}.

\subsubsection{High Pass filter for Image Sharpening}

High pass filters or a sharpening filter is used for preserving all the high-frequency information in an image while reducing low frequencies. A fraction of the image after passing it through a high-pass filter can be added to the original image to obtain an enhanced version of the input image \cite{b17}. However, high-pass filters are very sensitive to noise as depend mainly on elevating high frequencies and attenuating lower ones. 

However, Russo presents a new approach in \cite{b18} for the contrast enhancement of image based on a multiple output system. The chief advantage of this technique is the superior performance in the event of corruption due to Gaussian noise. This is done by adopting fuzzy models.

\subsubsection{Median filter for Image quality enhancement}

Median filters are order-statistics filters. These are a form of nonlinear smoothing operators used to perform noise reduction on an image or signal. Median filters are typically used for salt and pepper noise, also known as impulse noises, that can occur due to random bit error during image transmission or conversion \cite{b19}.

The median filtering algorithm works by running through a window of entries. This window slides over the entire signal. Suppose the window is of size $(2K+1)+(2L+1)$ at position $(k,l)$, then the input samples would be defined as – $u_{k-K,l-L},...,u_{k,l},...,u_{k+K,l+L}$.

\begin{figure}[htbp]
\centerline{\includegraphics[width=0.4\textwidth]{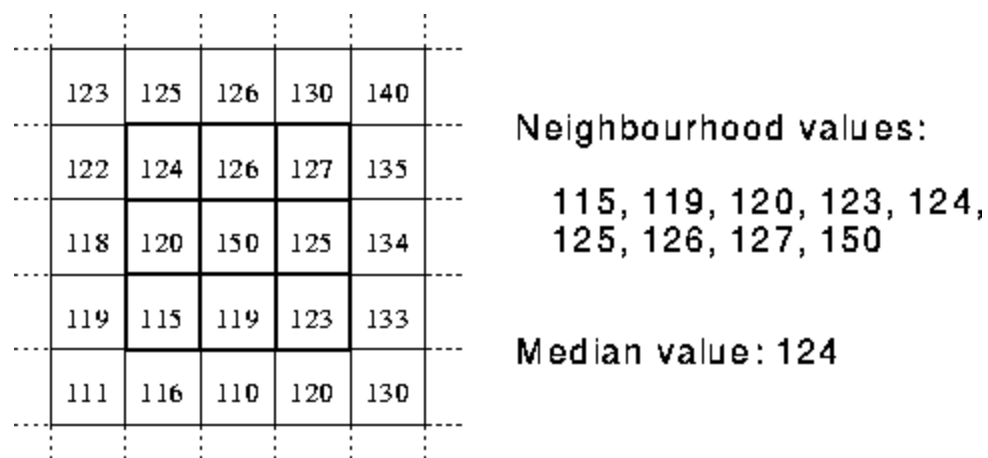}}
\caption{Calculation of median value using the neighborhood values\cite{b24}}
\label{fig5}
\end{figure}

Figure \ref{fig5} illustrates this calculation of the median value.

However, median filters present the issue of slight image blurring due as they also tend to smoothen the image details. To overcome this issue, Sun and Neuvo present a Detail-preserving median based filter in \cite{b20}. Their approach outperforms the weighted median filter \cite{b21}, stack filters \cite{b22} and adaptive weighted mean filer \cite{b23}. This approach removes impulses with minimal signal distortion while being detail preserving. Furthermore, unlike median filters, the detail-preserving median filter does not affect the image if impulse corruption is absent. Hence, making it an ideal prefilter for tumor extraction.

\subsection{Segmentation}

\subsubsection{Thresholding}

This is considered to be the most trivial method of image segmentation \cite{b25}. Equation \eqref{eq3} represents the thresholding process of converting a grayscale image into a binary image

\begin{equation}
g(x,y) = \begin{cases}
1 & \text{if} f(x,y) > T \\ 
0 & \text{if} f(x,y) \le T 
\end{cases}
\label{eq3}
\end{equation}

Where, $T$ is the fixed threshold value ranging between 0 and 255 and $g(x,y)$ is a binary intensity value (since it can only be 0 or 1) of a pixel at the spatial coordinate $(x,y)$.

\begin{figure}[htbp]
\centerline{\includegraphics[width=0.4\textwidth]{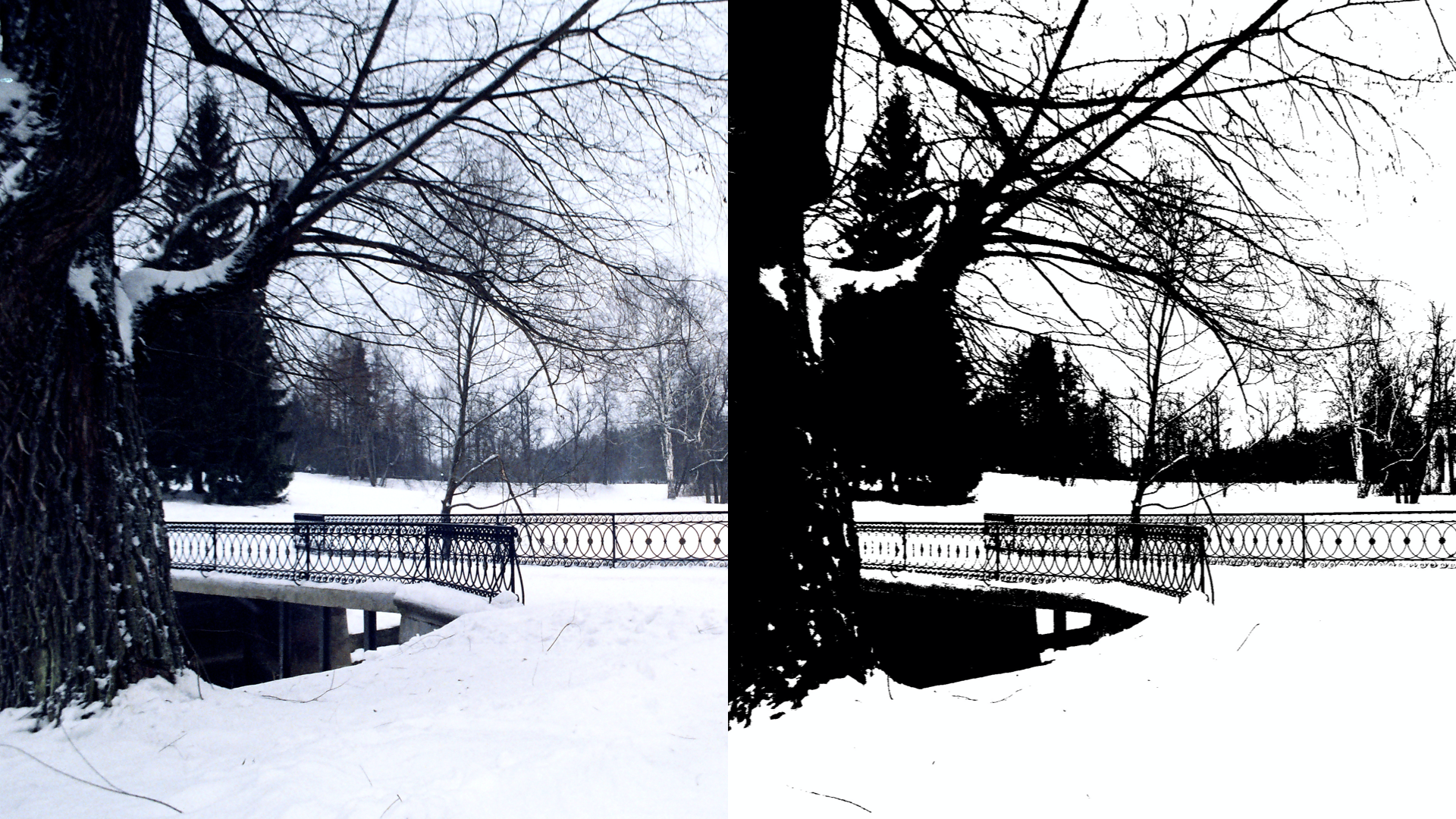}}
\caption{The effect of thresholding (\textit{right}) on an image (\textit{left}).\cite{b26}}
\label{fig6}
\end{figure}

Some of the common thresholding techniques are explained in \cite{b27}:

\subsubsection{Global Thresholding (Single Threshold)}

These are used when the differences between the foreground and background are very distinct. Have also proposed a novel global thresholding algorithm that uses boundary blocks for extracting a bimodal histogram \cite{b28}.

\begin{itemize}
\item{Traditional Thresholding (Otsu's method) \cite{b29} – used when the image has two distinct peaks in its histogram representation. This method calculates the optimum threshold separating the two classes such that their inter-class variance is maximum. }
\item{Iterative Thresholding (A new iterative triclass thresholding technique) \cite{b30}– This method first uses Otsu’s method to obtain the threshold and the means of the two separated classes. The image is then separated into three classes using the means derived from the two classes. The first two classes will not be processed further. They are termed as the foreground and the background. The third class is referred to as the ``To Be Determined" (TBD) region and is involved in the next iteration of triclass separation using Otsu’s method. This method identifies weak objects and reveals fine structures of complex objects better than Otsu’s original approach. }
\item{Multistage Thresholding (Quadratic Ratio Technique for Handwritten Character) – as the name suggests, QIR is used for retaining all the details of handwritten, hence it would not perform well for MRI images. Due to the use of fuzzy stage in the iteration, it performs better than other approaches for segmenting handwritten characters.}
\end{itemize}

\subsubsection{Local Thresholding}

\begin{itemize}
\item{Single Threshold – uses a single threshold value as described in equation \eqref{eq3}}
\item{Multiple Threshold \cite{b31} – segments image into multiple levels using its mean and variance.}
\end{itemize}

\begin{figure}[htbp]
\centerline{\includegraphics[width=0.4\textwidth]{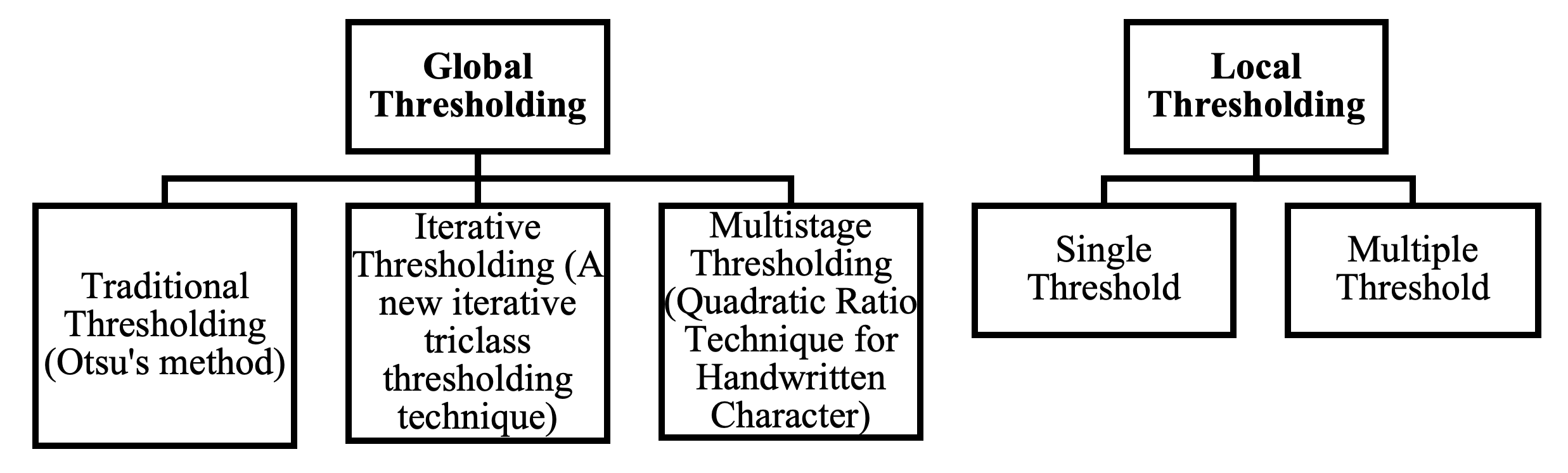}}
\caption{Some popular methods for image thresholding.\cite{b27}}
\label{fig7}
\end{figure}

Global thresholding methods tend to work well for medical images when the object of interest is significantly different from the background with respect to some characteristic. Such methods such as the one proposed by Bao and Zhang \cite{b32}, can also be used for noise detection while preserving edges in MRI images. Such methods also tend to perform better than wavelet-thresholding denoising methods. Furthermore, a multilevel thresholding method suggested by Manikandan et al. in \cite{b33} segments medical images by maximizing entropy. This method uses a real coded genetic algorithm with SBX crossover and performs more consistently for medical images.

\subsubsection{Watershed segmentation}

The watershed transformation process treats the gray-level image as a topographic relief. The brightness or intensity of each point is treated as its altitude. Based off of a geological watershed, a drop of water falls onto the surface, seeps along a path, then reaches a local minimum. This is used in the separation of adjacent drainage basins and find watershed lines.  Furthermore, as proposed by Najman and Schmitt in \cite{b35}, watershed algorithms can also be specified over a continuous domain.  Some of the different watershed definitions are:

\begin{itemize}
\item{Watershed by flooding – This method was proposed by Buecher and Lantuejoul in \cite{b36}. Their method extends the idea of drainage basins by continuously allowing ``water" from sources to collect in the local minima until the complete relief is flooded. Furthermore, a barrier is built where the ``water" sources meet. The arrangement of these barriers marks a watershed formed via flooding. One improvement of this method is the Priority-flood method \cite{b37}. }
\item{Watershed by topographic distance – This definition verifies that the catchment basin is the local minimum in the topographic relief.}
\item{Watershed by the drop of water principle – This idea was formally proposed by Cousty et al. in \cite{b38}. Intuitively the watershed of relief corresponds to the distinct local minima where a ``drop of water" can flow into.}
\end{itemize}

\subsubsection{Inter-pixel watershed algorithm}

This approach was proposed by Beucher and Meyer in \cite{b39}. The algorithm can be described as \ref{al1}.

\begin{algorithm}[htbp]
\SetAlgoLined
 \textbf{Initialize} a set \textit{S} with distinct label nodes for each minimum\;
 \For{\textit{S} $\neq \{\theta\}$}{
 \textbf{Extract} a node \textit{x} of minimum altitude\;
 \textbf{Attribute} the label of \textit{x} to each non-labeled node \textit{y} neighboring \textit{x}\;
\textbf{Insert} \textit{y} into set \textit{S}\;
 }
 \caption{Inter-pixel watershed algorithm}
 \label{al1}
\end{algorithm}

\subsubsection{Meyer’s flooding algorithm}

Proposed by Meyer and Maragos in \cite{b40}, this multiscale segmentation scheme works on grayscale images.

 A gradient image is used for the flooding process. Since successive flooding leads to the formation of adjacent catchment basins, basins emerge along the images.  
 
Hence the noise would lead to over-segmentation of the image. Thus, requiring that the data be preprocessed. Another approach is to merge regions based on similarity criterion afterwards. 

The algorithm works as described in \ref{al2}.

\begin{algorithm}[htbp]
\SetAlgoLined
\KwResult{non-labeled pixels as watershed lines}
\textbf{Initialize} a random set of seed markers for flooding, each with a distinct label\;
 \For{\textit{neighboring pixels} of each marker}{
 \textbf{Enqueue} pixel to priority queue \textbf{P}
(Associated priority is the gradient magnitude of each pixel)\;
 }
 \For{\textbf{P} $\ne \{\theta\}$}{
\textbf{Dequeue} pixel $p_l$ with least priority\;
\If{neighboring pixels of $p_l$  have the same labels}{
\textbf{label} $p_l$ with neighbors' label\;
}
\textbf{Enqueue} unlabeled neighboring pixels of $p_l$
 } 
 \caption{Meyer’s flooding algorithm}
 \label{al2}
\end{algorithm}

\subsubsection{Region Competition}

A novel algorithm proposed by Zhu and Yuille in \cite{b41} by unifies the following approaches:
\begin{itemize}
\item{snakes \cite{b42} and balloon methods \cite{b43}\cite{b44}\cite{b45}\cite{b46}}
\item{region growing and merging techniques \cite{b47}\cite{b48}\cite{b49}}
\item{Bayesian \cite{b50}\cite{b51}and Minimum Description Length (MDL) Criteria \cite{b52}\cite{b53}}
\end{itemize}

This multiband image segmentation technique is derived by minimizing a generalized Bayesian and MDL criterion. Furthermore, it also combines the statistical features of region growing and geometrical features of snakes and balloon methods.

An implementation by Amo et al. \cite{b54}, utilizes the region competition algorithm for road extraction from aerial images. The proposed implementation extracts roads, their centerlines, and sides. The algorithm utilizes the small changes in the curvature and radiometer of the road and its light appearance for extracting it from the aerial image. Hence, the implementation finds the region of interest, i.e., road margins, accurately and is robust. However, it requires a user to set seeds and hence is susceptible to human error \cite{b55}.

\subsection{Morphological Operations}

\begin{figure}[htbp]
\centerline{\includegraphics[width=0.3\textwidth]{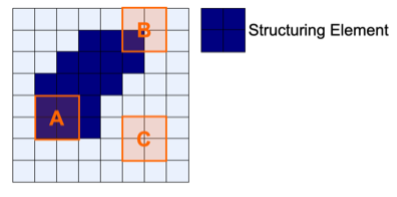}}
\caption{\textit{A} is a fit, \textit{B} is a hit, \textit{C} is neither a fit nor a hit, hence we term it as a miss. \cite{b55}}
\label{fig8}
\end{figure}

The last step may be morphological operations on the binary image formed. These are a collection of non-linear operations used to extract morphological features such as the form and structure of an image. Furthermore, morphological operations can also be used to remove imperfections in the segmented image. Morphological operations are performed using a structuring element to an input image, and the value is based on two factors. These are again illustrated in Figure \ref{fig8}.

\begin{itemize}
\item{\textbf{Fit}: All pixels on structuring element matches the pixels of the input image (\textit{A} in Fig.  \ref{fig8})}
\item{\textbf{Hit}: Any pixel on structuring element matches the pixels of the input image (\textit{B} in Fig.  \ref{fig8})}
\end{itemize}

Some of the basic morphological operations along with their equations can be found below.  Note that \textit{X} is the reference image and \textit{B} is the structuring element.

\begin{itemize}
\item{\textbf{Erosion}: Used for noise removal in the background and removal of holes in either the foreground or background. This process shrinks the foreground and enlarges the background. Given as: $X\ominus B = \{z|(B)_z \subseteq X \}$}
\item{\textbf{Dilation}: Enlarges the foreground and shrinks the background. Helps in enlarging the region of interest if it resides in the foreground. Furthermore, it is used for bridging gaps in an image since \textit{B} is expanding the features of \textit{X}. $X\oplus B = \{z| [(\hat{B})_z \cap X] \subseteq X \}$}
\item{\textbf{Opening}: Used to remove noise and Charged Coupled Defects (CCD) in images. This detail and simplifies images by rounding the corners from inside the object where the kernel fits. It is erosion followed by dilation.  $X\circ B= (X \ominus B) \oplus B$}
\item{\textbf{Closing}: Smoothens contours and maintains shapes and sizes of objects. Closing protects coarse structures, closes small gaps and rounds off concave corners. It is dilation followed by erosion. $X\bullet B = (X \oplus B) \ominus B$}
\end{itemize}

\subsection{Region filling method}

Region filling methods utilize morphological operations and are also termed as coloring. It is defined by equation \eqref{eq4}

\begin{equation}
X_k = \{X_{k-1} \oplus B\} \cap A^c, k=1,2,3...
\label{eq4}
\end{equation}
           
Where B denotes the structuring element, A denotes a set containing a subset whose elements are 8 connected boundary points of a region and k denotes the number of iterations. If the region is filled, then stop the iterations. A user could also predefine the number of iterations to fill the region.

Deb, Dutta and Roy propose a novel method for noise removal from brain images in \cite{b34}. This method uses region filling to denoise the image. Region filling takes place by interposing the pixel values from the boundaries of the region of interest. The method suggests the use of an interpolation method based on Laplace's equation to obtain the smoothest possible fills at the boundaries. However, this method requires user intervention to determine the region of interest. Furthermore, the selection of the region must be accurate.  

\section*{Acknowledgment}

The author, Jacob John would like to thank Dr. Prabu Sevugan for his continuous support throughout this paper. I would also like to thank Vellore Institute of Technology for their aid without which this paper wouldn't have been completed.

\end{document}